\documentclass[10pt,twocolumn,a4paper]{article}
\usepackage[margin=18mm,columnsep=6mm]{geometry}
\usepackage[T1]{fontenc}
\usepackage{mathptmx}
\usepackage{graphicx,booktabs,amsmath,amssymb,microtype,url}
\usepackage[hidelinks]{hyperref}
\usepackage{caption}
\usepackage{titlesec}
\titleformat{\section}{\large\bfseries}{\Roman{section}.}{0.5em}{}
\titleformat{\subsection}{\normalsize\bfseries}{\Alph{subsection}.}{0.5em}{}
\title{Evaluating Tibetan Speech Recognition Across Dialects and Adaptation Regimes: A Multi-Granularity Analysis}
\author{Jacob Moore$^{1}$, and Sheng Li$^{2}$\\[3pt]
\small $^{1}$Lawrence Technological University, Southfield, MI, USA\\
\small $^{2}$Institute of Science Tokyo, Yokohama, Japan\\
\small Corresponding authors: Jacob Moore and Sheng Li (\texttt{jmoore10@ltu.edu, sheng.li@ieee.org})}
\date{}
\begin{document}
\maketitle
\begin{abstract}
Tibetan speech recognition is usually evaluated with character or syllable error rates because the writing system does not mark word boundaries. These metrics can rank systems differently and respond differently to adaptation failures. We combine character error rate, syllable error rate, and segmented word error rate in one evaluation framework. We use dictionary-based, neural, and language-model-based segmentation and compare their boundaries with expert annotations. We evaluate five ASR architectures on Lhasa Tibetan, then test zero-shot transfer and target-corpus adaptation on five additional corpus subsets. We also vary LoRA rank and training-data size to examine adaptation failure. The results show that a lower character error does not always lead to a lower syllable or segmented-word error. These ranking differences persist across corpora. Metric correlations also weaken in some unsuccessful adaptation settings. Higher LoRA ranks improve syllable and segmented-word scores but remain worse than full fine-tuning in the tested range. A standardized phonological representation provides another view of spelling errors. These findings support evaluation at several linguistic levels and motivate better segmentation and adaptation methods for low-resource ASR.
\end{abstract}\\

\noindent\textbf{Index terms---}Tibetan speech recognition, cross-dialect transfer, segmented word error rate, low-resource languages, low-rank adaptation.

\section{Introduction}
Automatic speech recognition (ASR) converts speech into text. Evaluation should measure both spelling accuracy and accuracy at larger linguistic units. Tibetan marks syllable boundaries with the tsek, but does not consistently separate words with spaces. A word may contain one or several syllables. Computing word error rate (WER) therefore requires word segmentation.

Most Tibetan ASR studies report character error rate (CER). Some also report syllable error rate (SER) \cite{qin,zhao,liu}. CER is easy to calculate because it requires no word segmentation. However, a small character error can affect a whole syllable or multi-syllable word. Metrics at different levels can therefore favor models with different error patterns.

Our earlier IEEE Access study evaluated Tibetan ASR with segmented word error rate (SWER) \cite{original}. It applied Botok, BERT-UPOS, and Gemini segmentation before computing WER. Character-level and higher-level scores ranked models differently on NICT-Tib1. LoRA and quantized LoRA also performed poorly. That study used one Lhasa read-speech corpus and one adapter rank. It left two questions open. Do these metric differences occur in other corpora and dialects? Can larger adapters or more training data improve the poor adaptation results?

We address both questions in a common experimental design. We use the updated Lhasa reference and evaluate its five checkpoints on five additional corpus subsets. We also fine-tune each architecture on each target subset. The target data cover Amdo, Kham, Lhasa, and Shigatse speech. On NICT-Tib1, we test four LoRA ranks at four training-data fractions. A standardized phonological representation provides a separate view of spelling differences.

We study three sources of variation: architecture, corpus and dialect shift, and adapter capacity. Metric behavior changes across these settings. Syllable and segmented-word scores can reverse CER rankings. Their correlations with CER can also weaken after unsuccessful adaptation. Each SWER variant uses its own segmentation procedure, so its scores depend on the resulting token sequences.

The contributions are as follows.
\begin{enumerate}
\item We combine character-, syllable-, and segmented-word evaluation. We distinguish tokenization, score aggregation, and utterance-level correlation.
\item We extend a five-architecture Lhasa benchmark with 25 zero-shot transfer evaluations and 25 target-corpus adaptation runs across five additional corpus subsets.
\item We test 16 LoRA rank-by-data configurations and three new full-fine-tuning baselines under a fixed epoch budget.
\item We analyze model rankings, failure modes, and metric correlations. We use these findings to guide Tibetan ASR evaluation.
\end{enumerate}
The transfer experiments and rank ablation extend our published SWER study \cite{original}. Cross-corpus differences include recording conditions, speakers, text domains, and dialect.

\section{Background and Related Work}
\subsection{Tibetan units and segmentation}
Tibetan syllables can contain several orthographic components, including prefixes, superscripts, subscripts, vowel marks, and suffixes. Written forms can preserve distinctions that pronunciation reduces \cite{tournadre}. Short grammatical particles also carry important information. Evaluation at the character, syllable, and word levels, therefore, describes different aspects of the transcript.

Dictionary segmentation finds lexical entries in a sequence of syllables. Botok provides a public implementation \cite{botok}. It keeps an unrecognized syllable as a separate token. An ASR error can therefore break a dictionary match and split one word into several tokens. BoTok-SWER reflects both transcription errors and their effects on dictionary segmentation.

Statistical and neural segmenters use context to predict boundaries. Earlier methods include conditional random fields \cite{crf} and CNN-BiLSTM-CRF models \cite{cnn}. Tibetan transformer models also support contextual analysis \cite{tibert}. BERT-UPOS uses Tibetan BERT in a part-of-speech pipeline \cite{esupar}. We use its returned entities as evaluation units. These units may differ from linguistic word boundaries.

Segmentation also affects ASR evaluation in other writing systems. Ni and Leung showed that Chinese segmentation standards and algorithms change the reported WER \cite{ni}. A segmented-word score should therefore specify its segmenter and tokenization standard.

\subsection{Recognition architectures and adaptation}
Wav2Vec 2.0 and HuBERT learn speech representations through self-supervised pretraining \cite{wav2vec,hubert}. Their ASR models use connectionist temporal classification (CTC). CTC converts acoustic outputs into token sequences by collapsing repeats and removing blanks \cite{ctc}. Whisper uses an acoustic encoder and an autoregressive decoder trained on multilingual speech-text data \cite{whisper}. These systems differ in pretraining data, decoding objectives, and output representations. Their comparison, therefore, reflects complete systems, not decoder architecture alone.

LoRA introduces a low-rank update to a frozen weight matrix \cite{lora}. For a matrix $W_0$, the adapted weight is
\begin{equation}
W=W_0+\frac{\alpha}{r}BA,
\end{equation}
where $A$ and $B$ are trainable factors, and $r$ is the adapter rank. Quantized LoRA trains adapters on a quantized base model. QLoRA is a prominent implementation \cite{qlora}. Rank controls the available update space. Training data and optimization determine how the model uses that space. We test several ranks and data fractions to examine these factors.

\section{Evaluation Framework}
\subsection{Error rates at three granularities}
Let $R_i$ and $H_i$ denote the normalized reference and hypothesis for utterance $i$. For a tokenization function $T$, the normalized edit error is
\begin{equation}
E_i(T)=\frac{d(T(R_i),T(H_i))}{|T(R_i)|},
\end{equation}
where $d$ is the Levenshtein edit distance. CER uses character units; SER uses tsek-delimited syllables. For an automatic word segmenter $s$, we define
\begin{equation}
\mathrm{SWER}_s=\frac{S_s+D_s+I_s}{N_s},
\end{equation}
where the edits align $s(R)$ and $s(H)$, and $N_s$ is the reference-token count. Insertions can cause an error rate to exceed one, meaning that the edit count exceeds the reference length.

Botok produces BoTok-SWER, BERT-UPOS produces BERT-SWER, and Gemini 2.5 Flash Lite produces Gem-SWER. Both reference and hypothesis are segmented independently using the same method. BoTok-SWER and BERT-SWER cover the full test sets. Gem-SWER is available only for the original 500-utterance NICT-Tib1 subset; it was not computed for the cross-corpus or rank-ablation extensions.

The Gemini prompt requests pipe-delimited words and preserves the original Tibetan characters and tsek marks. Temperature is zero. The original study excluded per-utterance Gem-SWER values above 3.0 as segmentation failures. This threshold can also remove genuine ASR errors. Character-preservation checks provide a more direct way to detect invalid segmentation.

\subsection{Boundary agreement}
The original boundary study sampled 100 NICT-Tib1 test utterances, stratified by speaker. One expert annotated 2,634 inter-syllable positions. Each position received a binary label indicating whether a word boundary followed that syllable. Botok, Gemini, BERT-UPOS, and Claude Haiku segmented the same reference transcriptions independently. Claude served as an additional comparison and was not used to produce an ASR score.

We measure agreement with boundary F1 and Cohen's $\kappa$. Boundary rate is the fraction of inter-syllable positions marked as word boundaries. Gemini has the highest agreement with the expert (Table~\ref{tab:boundary}). Botok marks more boundaries and produces units closer to syllables. This study uses one annotator and clean Lhasa references. Boundary accuracy on ASR hypotheses and other dialects remains untested.

\begin{table}[t]
\centering\small
\caption{Boundary agreement on 100 NICT-Tib1 reference utterances, reproduced from \cite{original}. Rates are percentages.}\label{tab:boundary}
\begin{tabular}{lrrr}\toprule
Method & Boundary rate & $\kappa$ & F1\\\midrule
Human & 41.1 & --- & ---\\
Botok & 86.7 & 0.027 & 0.569\\
BERT-UPOS & 67.5 & 0.372 & 0.693\\
Gemini & 42.4 & 0.656 & 0.799\\
Claude Haiku & 44.3 & 0.333 & 0.617\\\bottomrule
\end{tabular}
\end{table}

BoTok-SWER measures errors under dictionary segmentation. Gem-SWER measures errors under LLM segmentation. Both depend on their tokenization rules. Neither directly measures listener comprehension or downstream task accuracy.

\subsection{Aggregation and uncertainty}
For $n$ utterances, let $e_i$ be the edit count and $N_i$ the reference-token count. We distinguish
\begin{align}
E_{\mathrm{micro}}&=\frac{\sum_i e_i}{\sum_i N_i},\\
E_{\mathrm{macro}}&=\frac{1}{n}\sum_i\frac{e_i}{N_i}.
\end{align}
Micro-averaging weights reference tokens equally. Macro-averaging weights utterances equally. Both can use utterance bootstrap sampling. Each micro bootstrap sample recomputes the ratio of total edits to total reference tokens. Each macro sample recomputes the mean utterance error.

We retain the reported macro-averaged confidence intervals (CIs), based on 10,000 utterance bootstrap samples. Primary corpus-level tables are labeled separately. Pearson correlations use per-utterance scores within each model-corpus cell unless stated otherwise. They describe how errors vary across utterances rather than how metrics rank different models.

Paired bootstrap differences assess model comparisons directly. Overlap between separate CIs does not establish equivalence. We report the extension's model differences descriptively because its reports do not include the raw predictions needed for new paired tests. With few test speakers, utterance-level CIs mainly describe uncertainty over the sampled utterances.

\subsection{Standardized phonological rendering}
We compute the standardized phonological error rate (PER) by mapping both transcripts through bophono's implementation of Tournadre's Manual of Standard Tibetan scheme, then calculating character edit errors. All corpora use the same mapping. This score shows how spelling differences change under a standard pronunciation representation. It does not measure dialect-specific pronunciations against manually annotated phones.

\section{Experimental Design}
\subsection{Corpora and splits}
Table~\ref{tab:data} summarizes the six corpus subsets. NICT-Tib1 provides the original Lhasa benchmark \cite{nict}. Its 20 speakers are split into 17 training speakers and three test speakers. The test set contains 1,547 utterances.

We use the official XBMU-AMDO31 split and custom speaker-disjoint splits for four TIBMD@MUC subsets \cite{xbmu,muc}. The subsets cover Amdo, Kham, Lhasa, and Shigatse. Their transcripts originally separate syllables with spaces. We replace these spaces with tsek marks to match the NICT-Tib1 evaluation convention.

\begin{table*}[t]
\centering\small
\caption{Corpus subsets and speaker-disjoint splits. MUC denotes TIBMD@MUC. Counts are training/test counts.}\label{tab:data}
\begin{tabular}{lllll}\toprule
Corpus & Dialect or variety & Utterances & Speakers & Split source\\\midrule
NICT-Tib1 & Lhasa/Central & 15,099 / 1,547 & 17 / 3 & Original study\\
XBMU-AMDO31 & Amdo & 18,539 / 2,041 & 52 / 6 & Official OpenSLR 133 split\\
MUC Amdo & Amdo & 18,820 / 4,134 & 19 / 4 & Custom speaker-disjoint split\\
MUC Kham & Kham & 4,005 / 429 & 10 / 2 & Custom speaker-disjoint split\\
MUC Lhasa & Lhasa & 26,830 / 3,519 & 40 / 8 & Custom speaker-disjoint split\\
MUC Shigatse & Central, distinct from Lhasa & 6,835 / 3,894 & 3 / 1 & Custom speaker-disjoint split\\\bottomrule
\end{tabular}
\end{table*}

MUC Shigatse has one test speaker, and Kham has two. The results therefore describe these corpus subsets, with limited coverage of speaker variation within each dialect.

\subsection{Architectures and optimization}
We compare Whisper Tiny, Base, and Small with Wav2Vec 2.0 Base and HuBERT Base. All target-corpus runs start from pretrained models.

The cross-corpus adaptation runs use 4,000 optimizer steps, 500 warmup steps, a per-device batch size of two, and eight gradient accumulation steps. The effective batch size is 16. Training uses fp16 and gradient checkpointing. Learning rates are $3.75\times10^{-5}$, $2.5\times10^{-5}$, and $1.25\times10^{-5}$ for Whisper Tiny, Base, and Small, respectively, and $10^{-4}$ for both CTC systems. CTC models use corpus-specific vocabularies with matching training and decoding blank/pad identifiers.

A fixed step budget gives different epoch counts across training sets. HuBERT completes 2.39 epochs on MUC Lhasa and 15.94 on MUC Kham. The protocol fixes optimizer updates, while data exposure varies with corpus size.

\subsection{Four complementary experiments}
\textit{Experiment 1: original Lhasa benchmark.} The earlier study compared five architectures. It also tested six settings for each Whisper size: standard fine-tuning, 8-bit and 4-bit post-training quantization, unquantized LoRA, and LoRA with 8-bit or 4-bit base weights. Adapters used rank eight, $\alpha=16$, and dropout 0.1. They targeted the key, query, value, and output projections, plus both feed-forward layers in every Whisper block. We retain the reported bit-width labels for these earlier quantized settings.

\textit{Experiment 2: zero-shot transfer.} We evaluate the five NICT-Tib1 checkpoints on each of the five additional test subsets. This gives 25 evaluations. The models have Tibetan training but no target-corpus adaptation.

\textit{Experiment 3: target-corpus adaptation.} We fine-tune each architecture on each target training subset using the fixed-step protocol. This gives 25 runs. Each transfer-adaptation comparison uses the same target test set.

\textit{Experiment 4: rank-by-data ablation.} We adapt Whisper Small on NICT-Tib1 with ranks $r\in\{4,8,16,32\}$ and data fractions $f\in\{25,50,75,100\}\%$. We sample utterances from the same 17 training speakers to preserve speaker diversity. Setting $\alpha=2r$ keeps $\alpha/r$ constant. Learned update norms can still differ across ranks.

Each of the 16 LoRA runs uses 4.24 epochs. This fixes the number of data passes, so smaller subsets receive fewer optimizer updates. Three full-fine-tuning runs use 25\%, 50\%, and 75\% of the data. The 100\% reference CER is 0.086; this condition was not rerun in the grid. We merge each trained adapter into the shared pretrained Whisper Small model before decoding. The grid tests unquantized LoRA.

\subsection{Gap analysis}
For each LoRA cell, we define the descriptive gap
\begin{equation}
G=\mathrm{BoTok\mbox{-}SWER}-\mathrm{CER}.
\end{equation}
The technical report fits $G$ to rank, data percentage, and their interaction using ordinary least squares:
\begin{equation}
G=\beta_0+\beta_r r+\beta_f f+\beta_{rf}rf+\epsilon,
\end{equation}
Here, $f$ is measured in percentage points. The gap compares errors with different token denominators, so we use it as a diagnostic measure. The regression uses 16 runs without repeated seeds. Its coefficient tests are exploratory.

\section{Results}
\subsection{Updated Lhasa reference}
Table~\ref{tab:reference} lists the updated NICT-Tib1 reference results. Wav2Vec2 has the lowest CER at 0.075, followed by Whisper Small at 0.086. CER and SER are strongly correlated within each model. The cross-corpus results below show whether they also favor the same models.

\begin{table*}[t]
\centering\small
\caption{Updated NICT-Tib1 reference from the cross-corpus report. Correlations are within-model, per-utterance Pearson correlations. The report does not separately label the aggregation of this reference CER column.}\label{tab:reference}
\begin{tabular}{lrrrrr}\toprule
Model & CER & $r$(CER,SER) & $r$(CER,BoTok) & $r$(SER,BoTok) & $r$(SER,BERT)\\\midrule
Whisper Tiny & 0.121 & 0.923 & 0.788 & 0.913 & 0.914\\
Whisper Base & 0.107 & 0.925 & 0.783 & 0.904 & 0.920\\
Whisper Small & 0.086 & 0.929 & 0.795 & 0.921 & 0.923\\
Wav2Vec2 Base & \textbf{0.075} & 0.886 & 0.839 & 0.868 & 0.921\\
HuBERT Base & 0.134 & 0.874 & 0.785 & 0.817 & 0.908\\\bottomrule
\end{tabular}
\end{table*}

Whisper Small's full-data full-fine-tuning reference CER is 0.086. At rank eight with full data, LoRA gives CER 0.714, SER 0.960, BoTok-SWER 0.978, BERT-SWER 0.951, and standardized PER 0.667. These values follow the updated cross-corpus and LoRA reports.

\subsection{Zero-shot transfer}
Table~\ref{tab:transfer} reports the 25 transfer evaluations. Error rates are high on the additional corpora, including MUC Lhasa. Transfer loss therefore includes corpus differences as well as dialect differences.

The ranking reversal remains clear on the Amdo subsets. On XBMU-AMDO31, Wav2Vec2 has the lowest CER at 0.472, while Whisper Small has lower SER (0.762 versus 0.869) and BoTok-SWER (1.014 versus 1.157). On MUC Amdo, Wav2Vec2 again has the lowest CER at 0.552, but Whisper Small performs better on SER, BoTok-SWER, and BERT-SWER. Character-level preservation therefore does not identify the best higher-granularity output in these transfer settings.

The same pattern appears within Central Tibetan. On MUC Lhasa, Wav2Vec2 has CER 0.382 versus Whisper Small's 0.520, but Whisper Small has lower SER and both segmented-word scores. On Shigatse, the corresponding CER values are 0.343 and 0.427, whereas BoTok-SWER favors Whisper Small at 0.797 versus 0.917. These comparisons show that the evaluation issue extends beyond the original test corpus.

\begin{table*}[t]
\centering\small
\caption{Zero-shot evaluation of NICT-Tib1-trained checkpoints. All values are reported as micro-averages from the cross-corpus report. W-Tiny/Base/Small denote Whisper models. Test counts are in Table~\ref{tab:data}.}\label{tab:transfer}
\begin{tabular}{llrrrr}\toprule
Corpus & Model & CER & SER & BoTok-SWER & BERT-SWER\\\midrule
XBMU Amdo & W-Tiny & 0.532 & 0.832 & 1.147 & 1.060 \\
 & W-Base & 0.500 & 0.793 & 1.104 & 1.030 \\
 & W-Small & 0.491 & 0.762 & 1.014 & 0.964 \\
 & Wav2Vec2 & 0.472 & 0.869 & 1.157 & 1.123 \\
 & HuBERT & 0.502 & 0.919 & 1.207 & 1.168 \\
\midrule
MUC Amdo & W-Tiny & 0.664 & 0.945 & 1.224 & 1.121 \\
 & W-Base & 0.642 & 0.924 & 1.201 & 1.107 \\
 & W-Small & 0.625 & 0.896 & 1.173 & 1.098 \\
 & Wav2Vec2 & 0.552 & 0.972 & 1.237 & 1.185 \\
 & HuBERT & 0.594 & 1.015 & 1.288 & 1.243 \\
\midrule
MUC Kham & W-Tiny & 0.626 & 0.918 & 1.233 & 1.119 \\
 & W-Base & 0.664 & 0.934 & 1.268 & 1.146 \\
 & W-Small & 0.596 & 0.873 & 1.181 & 1.104 \\
 & Wav2Vec2 & 0.514 & 0.898 & 1.184 & 1.130 \\
 & HuBERT & 0.543 & 0.927 & 1.195 & 1.163 \\
\midrule
MUC Lhasa & W-Tiny & 0.594 & 0.800 & 1.017 & 0.995 \\
 & W-Base & 0.541 & 0.744 & 0.945 & 0.952 \\
 & W-Small & 0.520 & 0.705 & 0.896 & 0.918 \\
 & Wav2Vec2 & 0.382 & 0.731 & 0.960 & 0.978 \\
 & HuBERT & 0.419 & 0.788 & 1.023 & 1.039 \\
\midrule
MUC Shigatse & W-Tiny & 0.510 & 0.704 & 0.934 & 0.937 \\
 & W-Base & 0.472 & 0.659 & 0.868 & 0.891 \\
 & W-Small & 0.427 & 0.601 & 0.797 & 0.840 \\
 & Wav2Vec2 & 0.343 & 0.680 & 0.917 & 0.972 \\
 & HuBERT & 0.392 & 0.757 & 1.011 & 1.043 \\
\bottomrule\end{tabular}
\end{table*}

\subsection{Target-corpus adaptation}
Table~\ref{tab:indomain} reports the 25 target-corpus training runs. Adaptation improves the four scores for every model on both Amdo subsets. For example, Whisper Small on XBMU-AMDO31 improves from CER 0.491 to 0.144 and from BoTok-SWER 1.014 to 0.295. On MUC Amdo, its BoTok-SWER falls from 1.173 to 0.292.

After adaptation, Wav2Vec2 still has the lowest CER on both Amdo subsets, while Whisper Small has the lowest SER and segmented-word scores. On XBMU-AMDO31, Wav2Vec2 achieves CER 0.105 compared with Whisper Small's 0.144; BoTok-SWER instead favors Whisper Small, 0.295 versus 0.353. On MUC Amdo, the difference is larger at the syllable and word levels: Whisper Small has SER 0.222 and BoTok-SWER 0.292, compared with 0.319 and 0.466 for Wav2Vec2.

Kham is more difficult under this protocol. All five models have higher CER after target-corpus adaptation than under transfer. However, the Whisper models and Wav2Vec2 have lower BoTok-SWER after adaptation. For Whisper Small, CER rises from 0.596 to 0.679, while BoTok-SWER falls from 1.181 to 1.078. The metrics therefore move in opposite directions on the same model and test set. Corpus size and training exposure vary together, which limits the interpretation of this result.

HuBERT produces empty predictions on MUC Kham and MUC Lhasa after adaptation. All four micro error rates equal one because every reference token is deleted. The lack of score variation also limits correlation analysis. These failures occur at both ends of the reported epoch range, so epoch exposure alone does not explain them.

\begin{table*}[t]
\centering\small
\caption{Standard fine-tuning on each target corpus under the fixed 4,000-step budget. All values are reported as micro-averages. HuBERT outputs are empty on MUC Kham and MUC Lhasa.}\label{tab:indomain}
\begin{tabular}{llrrrr}\toprule
Corpus & Model & CER & SER & BoTok-SWER & BERT-SWER\\\midrule
MUC Kham & W-Tiny & 0.693 & 0.930 & 1.015 & 0.983 \\
 & W-Base & 0.726 & 0.976 & 1.093 & 1.073 \\
 & W-Small & 0.679 & 0.931 & 1.078 & 1.041 \\
 & Wav2Vec2 & 0.603 & 0.894 & 0.990 & 0.963 \\
 & HuBERT & 1.000 & 1.000 & 1.000 & 1.000 \\
\midrule
XBMU Amdo & W-Tiny & 0.178 & 0.278 & 0.391 & 0.466 \\
 & W-Base & 0.162 & 0.249 & 0.345 & 0.426 \\
 & W-Small & 0.144 & 0.216 & 0.295 & 0.380 \\
 & Wav2Vec2 & 0.105 & 0.229 & 0.353 & 0.419 \\
 & HuBERT & 0.163 & 0.365 & 0.557 & 0.611 \\
\midrule
MUC Amdo & W-Tiny & 0.218 & 0.323 & 0.429 & 0.490 \\
 & W-Base & 0.189 & 0.275 & 0.366 & 0.430 \\
 & W-Small & 0.160 & 0.222 & 0.292 & 0.355 \\
 & Wav2Vec2 & 0.148 & 0.319 & 0.466 & 0.529 \\
 & HuBERT & 0.200 & 0.422 & 0.597 & 0.674 \\
\midrule
MUC Lhasa & W-Tiny & 0.373 & 0.506 & 0.614 & 0.669 \\
 & W-Base & 0.348 & 0.469 & 0.572 & 0.633 \\
 & W-Small & 0.315 & 0.410 & 0.498 & 0.571 \\
 & Wav2Vec2 & 0.233 & 0.468 & 0.630 & 0.689 \\
 & HuBERT & 1.000 & 1.000 & 1.000 & 1.000 \\
\midrule
MUC Shigatse & W-Tiny & 0.359 & 0.528 & 0.676 & 0.709 \\
 & W-Base & 0.332 & 0.485 & 0.628 & 0.664 \\
 & W-Small & 0.296 & 0.421 & 0.550 & 0.597 \\
 & Wav2Vec2 & 0.233 & 0.497 & 0.708 & 0.754 \\
 & HuBERT & 0.226 & 0.480 & 0.690 & 0.735 \\
\bottomrule\end{tabular}
\end{table*}

\subsection{Correlation changes across regimes}
We compute correlations within each model-corpus cell. On the NICT-Tib reference predictions, CER-SER correlations range from 0.874 to 0.929. For Whisper Small, CER-SER is 0.929, and SER-BoTok-SWER is 0.921. Utterances with more errors at one level generally have more errors at the other levels, even when the metrics favor different architectures.

CTC correlations weaken on Kham. Wav2Vec2's CER-BoTok-SWER correlation falls from 0.797 on XBMU-AMDO31 to 0.151 on Kham. CER-SER falls from 0.909 to 0.695, and BoTok-BERT falls from 0.804 to 0.044. The relation between metrics thus changes with the corpus and adaptation setting.

For Whisper, CER-SER correlations are at least 0.86 across transfer and target-adaptation cells. They are lower in the LoRA grid. Strong within-model correlations can therefore coexist with different architecture rankings. Poor adapter training can also weaken these correlations.

\subsection{Rank-by-data ablation}
Table~\ref{tab:ablation} reports the 19 new training configurations. Every LoRA cell remains far worse than the full-fine-tuning baselines. Increasing rank helps, especially at full data. From rank four to rank 32, SER falls from 0.958 to 0.919 and BoTok-SWER falls from 0.987 to 0.929. CER changes from 0.725 to 0.707. Greater capacity therefore improves all three measures, but the absolute error remains high.

More data also tends to improve performance. At rank 32, increasing the fraction from 25\% to 100\% reduces CER from 0.744 to 0.707 and BoTok-SWER from 0.972 to 0.929. At rank eight, CER improves with data while BoTok-SWER is lowest at 50\%, so the changes are not strictly monotonic across every metric and rank.

Full fine-tuning with 25\% of the data gives CER 0.455 and BoTok-SWER 0.641. Both are below the best LoRA values of 0.707 and 0.929. Full fine-tuning improves further at 50\% and 75\% data. Its full-data reference CER is 0.086. Thus, increasing rank and data within the tested LoRA grid does not recover full-fine-tuning performance.

\begin{table*}[t]
\centering\small
\caption{Whisper Small rank-by-data ablation on NICT-Tib1. The supplied report does not explicitly label the aggregation of this grid; values are preserved as reported. Standardized PER uses the common MST rendering. Full FT means full fine-tuning. The full-data reference provides CER only and was not rerun in the grid.}\label{tab:ablation}
\begin{tabular}{lrrrrrr}\toprule
Rank/method & Data (\%) & CER & SER & BoTok-SWER & BERT-SWER & Std. PER\\\midrule
4 & 25 & 0.751 & 1.007 & 1.060 & 1.055 & 0.741 \\
4 & 50 & 0.731 & 0.971 & 0.989 & 1.038 & 0.704 \\
4 & 75 & 0.730 & 0.964 & 0.981 & 0.956 & 0.694 \\
4 & 100 & 0.725 & 0.958 & 0.987 & 0.962 & 0.686 \\
8 & 25 & 0.752 & 0.986 & 1.034 & 1.032 & 0.740 \\
8 & 50 & 0.734 & 0.955 & 0.968 & 0.969 & 0.699 \\
8 & 75 & 0.715 & 0.951 & 0.972 & 0.961 & 0.673 \\
8 & 100 & 0.714 & 0.960 & 0.978 & 0.951 & 0.667 \\
16 & 25 & 0.744 & 0.975 & 0.994 & 0.983 & 0.717 \\
16 & 50 & 0.721 & 0.950 & 0.959 & 0.950 & 0.683 \\
16 & 75 & 0.712 & 0.949 & 0.961 & 0.949 & 0.671 \\
16 & 100 & 0.718 & 0.947 & 0.958 & 0.949 & 0.678 \\
32 & 25 & 0.744 & 0.960 & 0.972 & 0.957 & 0.713 \\
32 & 50 & 0.717 & 0.943 & 0.957 & 0.946 & 0.678 \\
32 & 75 & 0.715 & 0.931 & 0.944 & 0.933 & 0.669 \\
32 & 100 & 0.707 & 0.919 & 0.929 & 0.932 & 0.662 \\
full FT & 25 & 0.455 & 0.563 & 0.641 & 0.743 & 0.431 \\
full FT & 50 & 0.408 & 0.474 & 0.530 & 0.654 & 0.394 \\
full FT & 75 & 0.398 & 0.455 & 0.502 & 0.626 & 0.390 \\
full FT (reference) & 100 & 0.086 & --- & --- & --- & --- \\
\bottomrule\end{tabular}
\end{table*}

The gap regression gives a rank coefficient of $-0.00207$ ($p=0.0375$), a data coefficient of $-0.00040$ ($p=0.1223$), and an interaction coefficient of $0.000011$ ($p=0.3996$). The model has $R^2=0.596$. Because it includes an uncentered interaction, the rank coefficient applies at zero data. Within the tested range, the fitted rank slope is $-0.00207+0.000011f$. We interpret these estimates with the observed grid. The data coefficient's $p$ value does not rule out a data effect.

Mean per-cell CER-SER correlation increases from 0.489 at rank four to 0.663 at rank 32 (Table~\ref{tab:corr}). Mean CER-BoTok-SWER correlation rises from 0.233 to 0.336, with a non-monotonic intermediate value at rank eight. The full-fine-tuning subset models have much higher correlations. Increasing rank thus improves part of the correlation structure, but does not restore the full-fine-tuning behavior within this grid.

\begin{table}[t]\centering\small
\caption{Mean within-cell Pearson correlations in the ablation. Means average model-cell correlations, rather than pooling utterances.}\label{tab:corr}
\begin{tabular}{lrrr}\toprule
Group & Cells & CER-SER & CER-BoTok\\\midrule
Full FT & 3 & .955 & .846\\
LoRA rank 4 & 4 & .489 & .233\\
LoRA rank 8 & 4 & .516 & .190\\
LoRA rank 16 & 4 & .581 & .279\\
LoRA rank 32 & 4 & .663 & .336\\\bottomrule
\end{tabular}\end{table}

\begin{figure*}[t]
\centering\includegraphics[width=.96\textwidth]{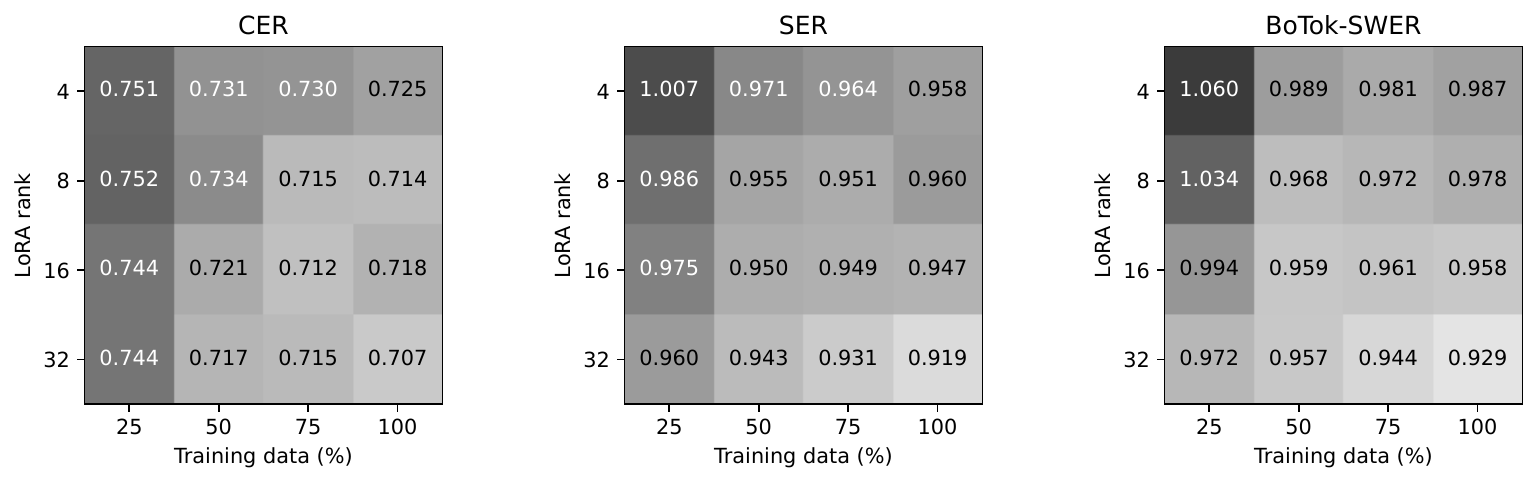}
\caption{The reported LoRA grid. Increasing capacity and data improves many cells, but all remain above the newly trained full-fine-tuning baselines in Table~\ref{tab:ablation}. Values and labels reproduce the extension report.}\label{fig:rank}
\end{figure*}

\subsection{Standardized phonological scores}
Standardized PER is often lower than orthographic CER in the macro results. The mean relative reduction is 19.1\% across 25 transfer cells and 13.4\% across 23 non-collapsed adaptation cells. These aggregate values follow the cross-corpus report.

For in-domain XBMU-AMDO31 Whisper Small, macro CER is 0.127 and standardized PER is 0.106. On Kham, the values are 0.699 and 0.616. The standard mapping removes some orthographic distinctions, but substantial error remains on Kham. Since all varieties use a Lhasa-based mapping, the scores describe standardized representations rather than actual dialect-specific pronunciation.

\section{Discussion}
\subsection{What segmented-word evaluation adds}
Two patterns emerge. First, the lowest CER often does not identify the lowest SER or SWER. This occurs on the original corpus and on additional Amdo and Central Tibetan subsets. Second, within-model correlations weaken in some difficult adaptation settings, especially the LoRA grid and Wav2Vec2 on Kham.

SER uses explicit syllable boundaries, so it requires no inferred word boundaries. In several comparisons, it selects the same model as SWER. SWER adds information about errors under a specific word segmentation procedure. Its value should therefore be assessed alongside SER as well as CER.

Botok can split a corrupted multi-syllable word into shorter tokens. This can increase the error count after segmentation. BERT-UPOS and Gemini use different boundaries and produce different score scales. Segment length, reference-token count, and hypothesis boundaries all affect SWER.

Gemini agrees best with the expert in the original Lhasa boundary study. The cross-corpus experiments use Botok and BERT-UPOS, whose boundaries were not validated on Amdo or Kham. Dialect-specific boundary annotations would help interpret their scores.

\subsection{What the ablation establishes}
Higher rank improves several LoRA scores, but all tested configurations remain worse than the new full-fine-tuning baselines. The tested increases in capacity and data therefore reduce errors without resolving the adaptation failure.

Several training choices may affect the remaining gap, including adapter placement, learning rate, update count, checkpoint selection, and decoding. The design fixes $\alpha/r$ and epoch count. It still allows learned update norms and optimizer update counts to vary. Repeated seeds and further optimization tests would help separate these effects.

The grid tests unquantized adapters merged into the base model before decoding. Extending its capacity findings to quantized LoRA requires a separate controlled comparison.

\subsection{Aggregation as part of the evaluation}
For a fixed tokenization and nonempty reference, micro error is a reference-length-weighted average of per-utterance error. The difference from macro error is
\begin{equation}
E_{\mathrm{micro}}-E_{\mathrm{macro}}=\frac{\operatorname{Cov}(N_i,E_i)}{\overline N}.
\end{equation}
A positive difference means that longer references tend to have higher utterance error rates. Short-reference errors receive more relative weight in the macro average.

For Whisper Tiny transferred to XBMU-AMDO31, micro BERT-SWER is 1.060 and macro BERT-SWER is 0.493. The gap shows how reference-length weighting changes the score. Repetition loops also occur in some predictions. Their contribution to the two aggregates requires per-utterance counts.

\subsection{Practical reporting recommendations}
Tibetan ASR evaluation should report CER and SER together. A named segmented-word score is also useful when word-level processing matters. Every table should state its aggregation. Micro and macro scores should use the same per-utterance edit counts.

SWER reporting should specify the segmenter version, configuration, normalization, and handling of malformed output. LLM segmentation also needs a fixed prompt and character-preservation checks. Boundary agreement should be assessed on both clean references and ASR hypotheses.

Model comparisons should use paired differences on the same test utterances. Correlation analyses should state whether they use individual cells, cell averages, or pooled predictions. Empty predictions and repetition loops should be counted separately to identify failure types.

\subsection{Scope and reproducibility}
The original study provides a public evaluation package and a 100-utterance boundary dataset. This study uses the updated cross-corpus and LoRA reference scores, correlations, and regression results. Raw per-utterance counts would allow further paired analyses.

Reproduction requires model identifiers and revisions, speaker splits, random seeds, and data-sampling lists. It also requires optimizer settings, adapter targets, checkpoint selection, decoding arguments, and segmenter versions. Released scoring records should include per-utterance edit counts and separate micro and macro aggregates. Quantization experiments should specify the bit-width implementation.

The experiments cover read speech with limited test-speaker diversity. Future work should evaluate conversational speech, listener comprehension, and downstream translation or retrieval. Multiple boundary annotators and dialect-specific phonological mappings would also improve the evaluation.

\section{Conclusion}
We evaluated Tibetan ASR across architectures, corpora, and adaptation settings. Lower character error did not consistently lead to lower syllable or segmented-word error. Several ranking reversals persisted across corpora. Metric correlations also changed under difficult adaptation conditions. Higher LoRA ranks improved some scores but remained worse than full fine-tuning within the tested grid. Standardized phonological scores provided another view of spelling errors. These results support evaluation at several linguistic levels, with explicit choices for segmentation and aggregation. Future work should link these scores to dialect-aware human judgments and downstream performance and improve adaptation under limited supervision.

%\section*{Acknowledgment}
%This work was supported by JST BOOST, Japan, under Grant JPMJBY25F6.
\bibliographystyle{unsrt}
\bibliography{references}
\end{document}